\documentclass[final,5p,times,twocolu]{elsarticle}
\usepackage{amssymb}
\usepackage{amsmath}
\usepackage{lineno}

\makeatletter
\def\ps@pprintTitle{%
 \let\@oddhead\@empty
 \let\@evenhead\@empty
 \def\@oddfoot{}%
 \let\@evenfoot\@oddfoot}
\makeatother

\usepackage{algorithm}
\usepackage[noend]{algpseudocode}

\makeatletter
\def\BState{\State\hskip-\ALG@thistlm}
\makeatother
\usepackage{pdfpages}

\usepackage{textcomp}

\usepackage{graphicx}
\usepackage{longtable}
\usepackage{array}

\newcolumntype{C}[1]{>{\centering\let\newline\\\arraybackslash\hspace{0pt}}m{#1}}

\usepackage{caption}
\usepackage{subcaption}

\usepackage{float}

\usepackage{multicol}

\begin{document}

\begin{frontmatter}

%% Title, authors and addresses

%% use the tnoteref command within \title for footnotes;
%% use the tnotetext command for the associated footnote;
%% use the fnref command within \author or \address for footnotes;
%% use the fntext command for the associated footnote;
%% use the corref command within \author for corresponding author footnotes;
%% use the cortext command for the associated footnote;
%% use the ead command for the email address,
%% and the form \ead[url] for the home page:
%%
%% \title{Title\tnoteref{label1}}
%% \tnotetext[label1]{}
%% \author{Name\corref{cor1}\fnref{label2}}
%% \ead{email address}
%% \ead[url]{home page}
%% \fntext[label2]{}
%% \cortext[cor1]{}
%% \address{Address\fnref{label3}}
%% \fntext[label3]{}

\title{\textbf{Bayesian Optimization for Parameter Tuning of the XOR Neural Network}}

%% use optional labels to link authors explicitly to addresses:
%% \author[label1,label2]{<author name>}
%% \address[label1]{<address>}
%% \address[label2]{<address>}

\author[]{ $\dag$ $\textasteriskcentered$ L. Stewart }
\ead{ls3914@ic.ac.uk}
\author[]{$\textasteriskcentered$ M. Stalzer }
\ead{stalzer@caltech.edu}

\address{ $\dag$ Imperial College London}
\address{$\textasteriskcentered$ California Institute of Technology}
\begin{abstract}
%% Text of abstract  %%ABSTRACT SHOULD BE ONE PARAGRAPH
%%hint at conclusion at end of abstract with a little bit of results.
When applying Machine Learning techniques to problems, one must select model parameters to ensure that the system converges but also does not become stuck at the objective function's local minimum.  Tuning these parameters becomes a non-trivial task for large models and it is not always apparent if the user has found the optimal parameters.  We aim to automate the process of tuning a Neural Network, (where only a limited number of parameter search attempts are available)  by implementing Bayesian Optimization. In particular, by assigning Gaussian Process Priors to the parameter space, we utilize Bayesian Optimization to tune an Artificial Neural Network used to learn the XOR function, with the result of achieving higher prediction accuracy. \\

\end{abstract}

%\begin{keyword}
%Science \sep Publication \sep Complicated
%% keywords here, in the form: keyword \sep keyword

%% MSC codes here, in the form: \MSC code \sep code
%% or \MSC[2008] code \sep code (2000 is the default)

%\end{keyword}

\end{frontmatter}

%%
%% Start line numbering here if you want
%%
%\linenumbers

%% main text
\section{Introduction}
\label{S:1}

\noindent This paper addresses the use of Bayesian Optimization for parameter tuning in Artificial Neural Networks that learn by Gradient Descent. We will focus on what  are arguably the two most important network parameters, the learning rate $\eta$, and the node activation function hyper-parameter $\theta$. For clarification on the definition of $\theta$, suppose node $k$ in a Neural Network is a hyperbolic tanh node with hyper-parameter $\theta$. Then the output of this node, $\mathcal{O}_{k,\theta}$ will be: 

$$ \mathcal{O}_{k,\theta}(\textbf{x})=\frac{e^{ \, \theta \sum_{1}^{n}w_ix_i}-e^{- \, \theta \sum_{1}^{n}w_ix_i}}{e^{\, \theta \sum_{1}^{n}w_ix_i}+e^{\,- \theta \sum_{1}^{n}w_ix_i}}$$

where $\textbf{x}=\begin{pmatrix}
x_1 & x_2 & \cdots & x_n
\end{pmatrix}
\in \mathbb{R}^n$ and $w_i$ are the synaptic weighting elements ( i.e. $\theta$ is the scale factor for the node function.) Whilst we have opted with $\alpha$ and $\theta$ as our parameters, the ideas discussed throughout this paper generalise to any number of chosen parameters.\\

The selection of such parameters can be critical to a Network's learning process. For example, a learning rate $\eta$ that is too small can stop the system from converging, whilst taking $\eta$ too high could lead to the system becoming stuck in a local minimum of the cost function. It can often be computationally expensive or time costly to evaluate a large set of network parameters and in this case finding the correct tuning becomes an art in itself. We consider an automated parameter tuning technique for scenarios in which an exhaustive parameter search is too costly to evaluate.\\

\subsection{Translation to a Black-Box Problem}

\noindent Suppose we have a fixed Neural Network topology with $\xi$ epochs of a chosen learning method (e.g 10,000 epochs of Gradient Descent Back-Propagation). Then we denote this by $N_{\xi}$. We also denote the network parameter space (the set of free parameters available to tune the network) for $N_{\xi}$ by $\mathbb{P}$. For example, if we wish to tune the learning rate $\alpha \in (0.001,1)$ and the activation function hyper-parameter $\theta \in (0.001,1)$, then $\mathbb{P}=(0.001,1)\times (0.001,1) \subset \mathbb{R}^2$. \\

With this terminology one can view the objective/cost function that is associated with the Neural Network as a function of the parameter space for $N_{\xi}$:
\begin{equation}
\Gamma_{N_{\xi}}: \mathbb{P} \longrightarrow \mathbb{R}
\end{equation}
The objective to optimally tune a network $N_{\xi}$ over a parameter space $\mathbb{P}$ can now be summarised as finding $\textbf{p} ^* \in \mathbb{P}$ such that:

\begin{equation}
\textbf{p}^*=\underset{\textbf{p} \in \mathbb{P}}{\text{argmin}} \, \, \Gamma_{N_{\xi}}
\end{equation}

\subsection{Bayesian Optimization}
\noindent Bayesian Optimization is a powerful algorithmic approach to finding the extrema of Black-Box functions that are costly to evaluate. Bayesian Optimization techniques are some of the most efficient approaches with respect to the number of function evaluations required \cite{mockus1994application} .  Suppose we have $k$ observations $\mathcal{D}_{1:k}= \{ \, \textbf{x}_{1:k}, \,  \Gamma_{N_{\xi}} (\textbf{x}_{1:k}) \,  \} $ where $\textbf{x}_i \in \mathbb{P}$. The Bayesian Optimization algorithm works by using a Utility function $\mathcal{U} : \mathbb{P} \longrightarrow \mathbb{R}$ in order to decide the next best point to evaluate. The algorithm written in terms of the parameter optimization problem (2) can be summarised as follows:
  \medskip
\begin{algorithm}[H]
\caption{Bayesian Optimization}\label{euclid}
\begin{algorithmic}[1]
\For{$i =1$ to $n$}
            \State Find $\textbf{x}^*=\underset{\textbf{x} \in \mathbb{P}}{\text{argmin}} \, \mathcal{U}(\textbf{x} \, | \, \mathcal{D}_{1:k} \, )$ 
            \State Sample the objective function at the chosen point  $ \Gamma_{N_{\xi}} (\textbf{x}^*)$
            \State Augment the data $\mathcal{
            D}_{1:k+1}=\mathcal{D}_{1:k}\cup \Big \{ \big (\, \textbf{x}^*,\Gamma_{N_{\xi}} (\textbf{x}^*)  \, \big) \Big \} $
            \State Update the Prior Distribution
\EndFor           

\end{algorithmic}
\end{algorithm}
\medskip

The algorithm terminates when $i$ reaches $n$ or if the Utility Function chooses the same element of $\mathbb{P}$ consecutively.  For more information on Bayesian Optimization see \cite{nandotut}.

\subsection{Choice of Parameter Space, Prior and Utility Function}

\noindent This paper aims to optimize learning rate $\alpha$ and node function hyper-parameter $\theta$ as in \textit{Section (1.3)} \,  i.e.  $\mathbb{P}=(0.001,1) \times (0.001,1)$. We will take our prior distribution  (as described in Algorithm 1) to be a Gaussian Process with a standard square exponential kernal \citep{rasmussen_gaussian}: 
$$k (\, \textbf{x}_i, \textbf{x}_j \, )=exp \, \Big ( -\frac{1}{2} ||\textbf{x}_i-\textbf{x}_j ||^2 \, \Big )$$ 

\noindent For speed the Julia code does not run the niave $\mathcal{O} (n^3)$ Sherman-Morrison-Woodbury matrix multiplication \cite{rasmussen_gaussian}. Instead it uses a Cholesky method detailed in \cite{rasmussen_gaussian}; both methods are explained further within the code file. For very large parameter spaces one may want to look into using a Gaussian Process update method of $\mathcal{O}\big (n log^2(n) \big)$ \cite{ambikasaran2014fast}. \\

The chosen utility function is the Lower Confidence Bound \cite{nandotut}: $$\text{LCB} \,  ( \textbf{x}_i)=\mu ( \textbf{x}_i) + \gamma \sigma ( \textbf{x}_i) \quad \gamma \in (0,1)$$ It is worth noting that whilst this utility function is indeed sufficient for tuning the XOR Neural Network, it may be necessary to use an IMGPO utility function (exponential convergence) for larger scale problems   \citep{kawaguchi2015bayesian}.

\section{XOR Neural Network Optimization}

\noindent Consider the XOR function \cite{compbook} (Exclusive Or) defined as:
\begin{align*}
\text{XOR} : \mathbb{Z}_2^{\oplus 2} &\rightarrow \mathbb{Z}_2 \\
(x,y)& \mapsto(x+y) \, \text{mod}(2)
\end{align*}
XOR has traditionally been a function of interest as it cannot be learned by a simple 2 layer Neural Network \citep{elman1990finding}. Instead the problem requires a 3 layer Neural Network, (hence there will be 3 nodes affected by activation function hyper-parameter tuning.) Figure 1 shows the topology of the Neural Network required to learn the XOR function. The Neural Network consists of an input layer followed by two hidden layers of sigmoid functions, where the output of a sigmoid node is $\mathcal{O}_\theta (\textbf{x}) =\frac{1}{1+e^{-\theta \sum_1^n w_i x_i}}$.  \\

\begin{center}
\textbf{XOR Neural Network}\\

\begin{figure}[H]
\centering
	\includegraphics[scale=0.25]{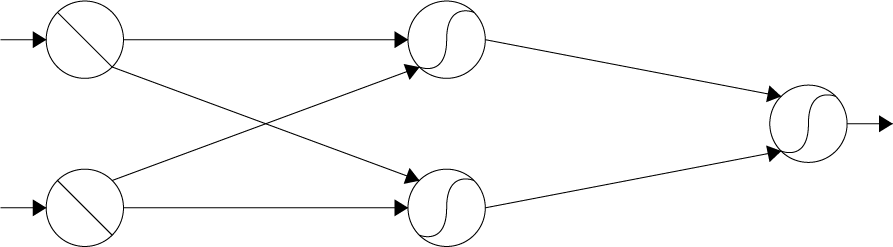}
\end{figure}
 \captionof{figure}{}
\end{center}
 \medskip
The target goal is to optimize $\alpha$, $\theta$ over the space $\mathbb{P}=(0.001,1)\times (0.001,1)$. The Neural Network's attributed loss function is the Mean Square Error, the network's topology $X$ to be as shown in Figure 1 and also $\xi=1000$. This can all be neatly summarised as $\Gamma_{N_{\xi}}=\text{MSE}_{X_{ \, 1000}}$. \\

For Deep Neural Networks or any Neural Network where the number of epochs $\xi$ is large, it can become very time costly to evaluate  $\text{MSE}_{X_{ \, \xi}} \Big [ (\alpha_i,\theta_j) \Big]$. Furthermore,  to achieve a small step size when discretizing $\mathbb{P}$ an exhaustive search of the Network parameters becomes time costly. In fact, for higher dimensional variants of $\mathbb{P}$  it becomes simply impossible to exhaustively evaluate the parameter space. To keep our proposed solution generalised we will limit the number of parameter searches allowed to 20. \\

\subsection{Code Structure}
%%% Here is image placement for next page as it is all messed up if
%%you do it on the actual page
\begin{figure*}[h!] %[H] for inline
\centering
\begin{minipage}{.5\textwidth}
  \centering
\includegraphics[height=5cm]{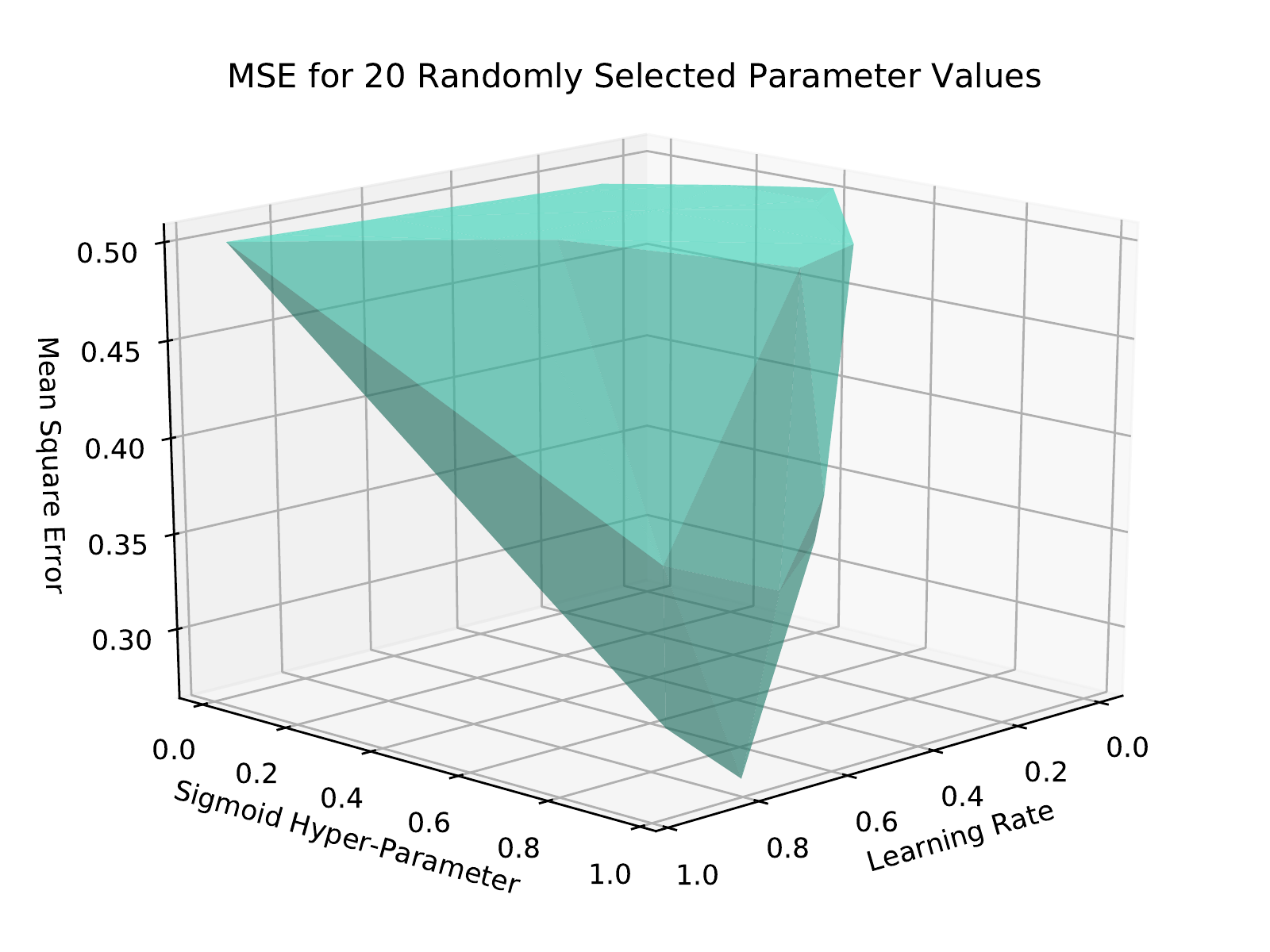}
%\addtocounter{figure}{+1}
  \captionof{figure}{}
  \label{fig:test1}
\end{minipage}%
\begin{minipage}{.5\textwidth}
 \centering
\includegraphics[height=5cm]{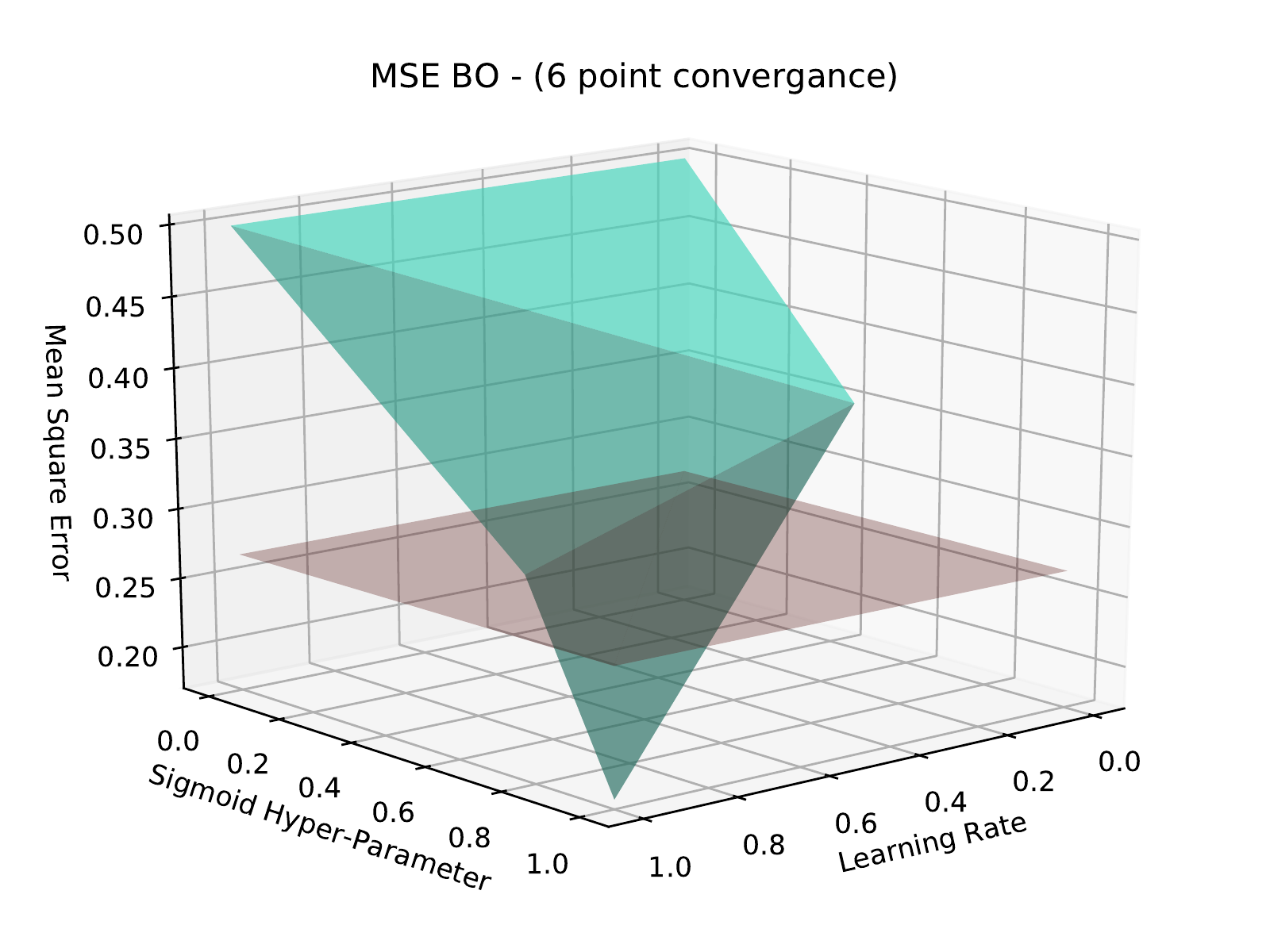}
 \captionof{figure}{}
 \label{fig:test2}
\end{minipage}
\end{figure*}

\noindent The Julia code is available with this pre-print on arXiv. The code is distributed under a Creative Commons Attribution 4.0 International Public License. If you use this code please attribute to L. Stewart and M.A. Stalzer  \textit{Bayesian Optimization for Parameter Tuning of the XOR Neural Network}, 2017.\\

The code compares Bayesian Optimization to a Random Grid Search of $\mathbb{P}$, a technique commonly used to achieve an acceptable tuning of Neural Networks. During the Random Grid Search process the code selects 20 random points of $\mathbb{P}$ and evaluates $\text{MSE}_{X_{ \, 1000}}$ for each point. During Bayesian Optimization, the code computes $\text{MSE}_{X_{ \, 1000}}$ for one random element of $\mathbb{P}$ say $\textbf{p}_1$ and forms $\mathcal{D}_1=\big \{ (\textbf{p}_1, \text{MSE}_{X_{ \, 1000}} (\textbf{p}_1) \big \}$. It then runs the Bayesian Optimization algorithm described in \textit{Section 1.2} where $\gamma \equiv 1$. 

\section{Results}
\noindent The results for the experiment denoted above in Section 2 are as follows:

\begin{table}[H]
\begin{center}
%add a title to the table here saying table 1
\medskip
\begin{tabular}{| C{2.2cm} | C{2.2cm} | C{2.2cm} |}
\hline
&  Random Search & Bayesian Opt  \\
\hline
$\text{MSE}_{X_{ \, 1000}}$ &   0.2681 & 0.1767\\
\hline 
\# Search Points & 20  & 6\\
\hline
Run Time $(s)$ & 1.5411 &  0.6012 \\
\hline
\end{tabular}
\end{center}
\caption{}
\end{table}

Where $\#$ Search Points is the number of points that each of the processes evaluated (this will always be 20 for Random Search as it has no convergence criterion). Figure 2 shows the Mean Square Error plot for the $20$ randomly selected elements of $\mathbb{P}$, whilst Figure 3 shows the Mean Square Error plot for the points selected by Bayesian Optimization. The Merlot colored plane in Figure 3 denotes the minimum MSE reached by Random Grid Search for comparative purposes. \\

Figure 4 shows the development of Mean Square Error with time for Bayesian Optimization and also includes Random Grid Search for ease of error comparison.  \\

\begin{center}
\includegraphics[height=4.75cm]{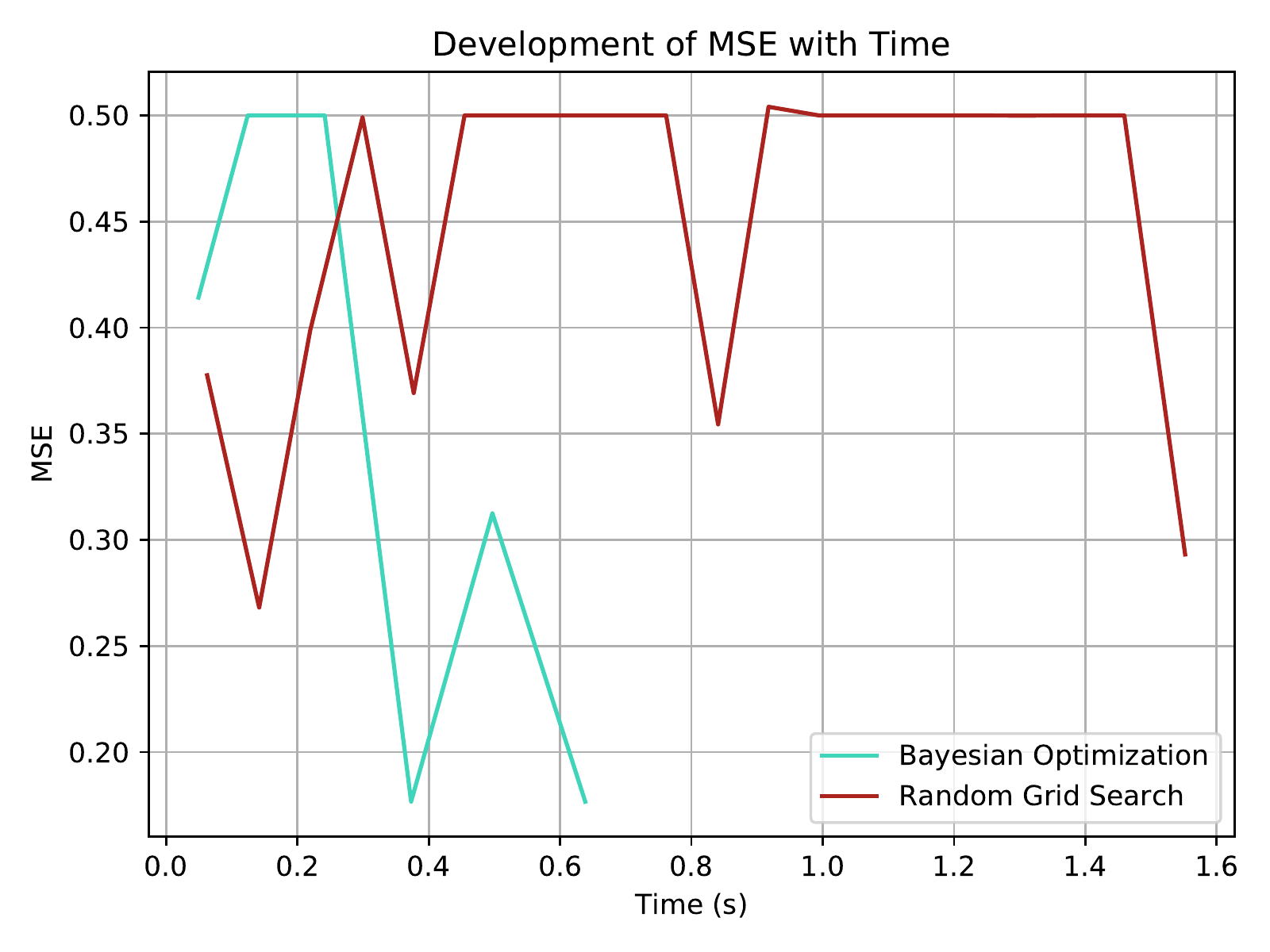} 
\captionof{figure}{}
\end{center}

It is clear Bayesian Optimization achieved higher accuracy whilst also running faster than Random Grid Search. This is due to the fact that the algorithm converged to the optimal value with respect to the LCB Metric.\\

This raises the question of how much more accurate is the result obtained by the LCB Metric (i.e.\,is the LCB a suitable Metric for tuning the XOR Neural Network)? To answer this it is useful to see the time required for Random Grid Search to achieve set a selection of threshold Mean Square Errors, including the Bayesian Optimal 0.1767. The results for this are displayed in Table 2. \\

\begin{table}
\begin{center}
\begin{tabular}{| C{2cm} | C{2cm} | C{2cm} |}

\hline
Threshold MSE &  \#Search Points & Time to Reach Threshold (s)  \\
\hline
0.190 &   2034 & 121.8\\
\hline 
0.185 & 5685  & 454.5\\
\hline
0.177 & N/A &  N/A \\
\hline
\end{tabular}
\end{center}
\caption{}
\end{table}

%When setting the threshold value to be the 3 decimal place round up of the Bayesian Optimal (0.177)
\par

One can see that for Random Grid Search to reach a moderately lower error of just 0.190, the consequential time cost vastly increases. When setting the threshold value to the Bayesian Optimal (0.177), Random Grid Search did not even converge.  
\section{Concluding Remarks and Future Works}

\noindent We have shown in Section 3 that Bayesian Optimization can be successfully implemented to attain a highly accurate tuning of the XOR Neural Network in a very fast time. There are many avenues for future development, a few of which are briefly summarised below:

\begin{itemize}
\item Apply Bayesian Optimization to more complex Neural Network topologies i.e. Recurrent or Deep Neural Nets. To do this a more sophisticated utility function may be required. \cite{kawaguchi2015bayesian}.
\item Apply to Bayesian Optimization to tuning higher dimensional variants of $\mathbb{P}$. Further parameters could be added e.g. momentum \cite{momentum} and regularization rate \cite{regular}. Another possibility is to have a different learning rate/node function hyper-parameter for each synapse/node. For both these tasks the use of Random Embeddings may be useful \cite{wang2016bayesian}.
\item Test a wide variety of Gaussian Process Kernals and use a faster Kernal update method \cite{ambikasaran2014fast}.  It would also be interesting to investigate the effect of using different prior distributions in Bayesian Optimization \cite{DNGO}.
\end{itemize}

\newpage
\onecolumn
\section*{Acknowledgements}
This research is funded by the Caltech SURF program and the Gordon and Betty Moore Foundation through Grant GBMF4915 to the Caltech Center for Data-Driven Discovery.
% \newpage

%% The Appendices part is started with the command \appendix;
%% appendix sections are then done as normal sections
%% \appendix

%\section*{References}

%% \label{}

%% References
%%
%% Following citation commands can be used in the body text:
%% Usage of \cite is as follows:
%%   \cite{key}          ==>>  [#]
%%   \cite[chap. 2]{key} ==>>  [#, chap. 2]
%%   \citet{key}         ==>>  Author [#]

%% References with bibTeX database:

\bibliographystyle{unsrt}
\bibliography{References}

%\bibliographystyle{model1-num-names}

%% Authors are advised to submit their bibtex database files. They are
%% requested to list a bibtex style file in the manuscript if they do
%% not want to use model1-num-names.bst.

%% References without bibTeX database:

% \begin{thebibliography}{00}

%% \bibitem must have the following form:
%%   \bibitem{key}...
%%

% \bibitem{}

% \end{thebibliography}

\end{document}